\documentclass[runningheads]{llncs}
\usepackage[T1]{fontenc}
\usepackage{graphicx,verbatim}
\usepackage{booktabs}
\usepackage{multirow}
\usepackage{siunitx} 
\usepackage{amssymb}
\usepackage{amsmath}
\usepackage{bm}
\usepackage[breaklinks=true]{hyperref} 
\PassOptionsToPackage{breaklinks=true}{hyperref}
\usepackage{orcidlink}
\begin{document}

\newcommand\figexplainerillposed{
\begin{figure}[t]
    \includegraphics[width=1\linewidth]{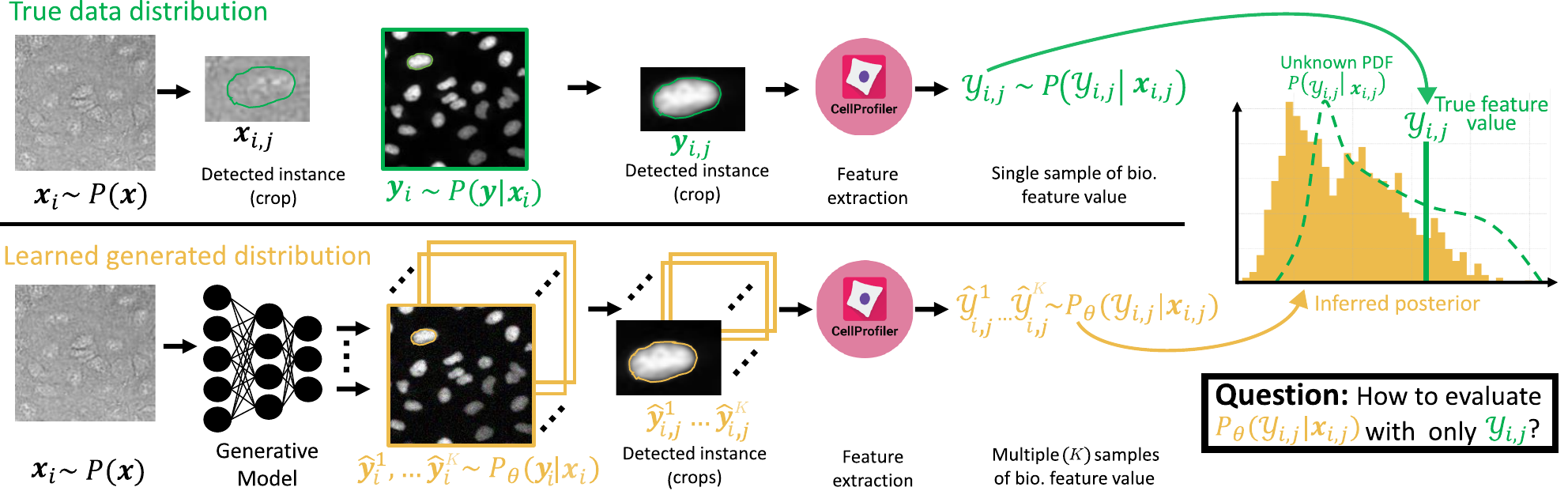}
    \caption{\textbf{How to evaluate a predicted posterior distribution with a single target sample? }A traditional HTS pipeline (top) uses fluorescence microscopy and extracts feature values $\mathcal{Y}_{i,j}$ for each cell $j$ in image $i$.
    A cell's true feature value $\mathcal{Y}_{i,j}$ can be seen as a single sample from an inaccessible posterior $P(\mathcal{Y}_{i,j}|\textbf{x}_{i,j})$.
    Although a generative VS model can produce many samples (bottom) $\hat{\mathcal{Y}}_{i,j}^{1,\ldots,K}$, its learned distribution must be evaluated against the single true value.}
   \label{fig:figexplainerillposed}
\end{figure}
}

\newcommand\metricsdd{
\begin{figure}[t]
    \includegraphics[width=1\linewidth]{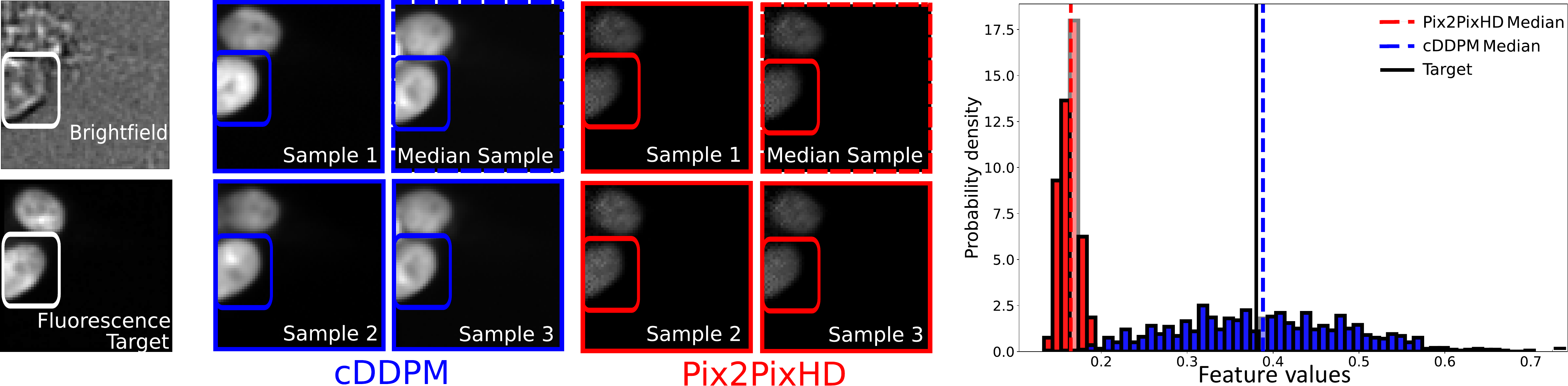}
    \caption{
    \textbf{Qualitative results for single input image}. 
    For an unseen bright-field, we show three model samples and the sample corresponding to the median posterior result. Alongside, we display the predicted posterior distribution for the 'upper quartile intensity' feature (F7), highlighting the median values and the target feature value. 
    Pix2PixHD underestimates cell intensity, while cDDPM exhibits greater variability; its posterior is wider and closer to the target value.
    }
    \label{fig:metrics_deepdive}
    \end{figure}
}

\newcommand\figcmodels{
    \begin{figure}[t]
        \centering
        \includegraphics[width=\textwidth]{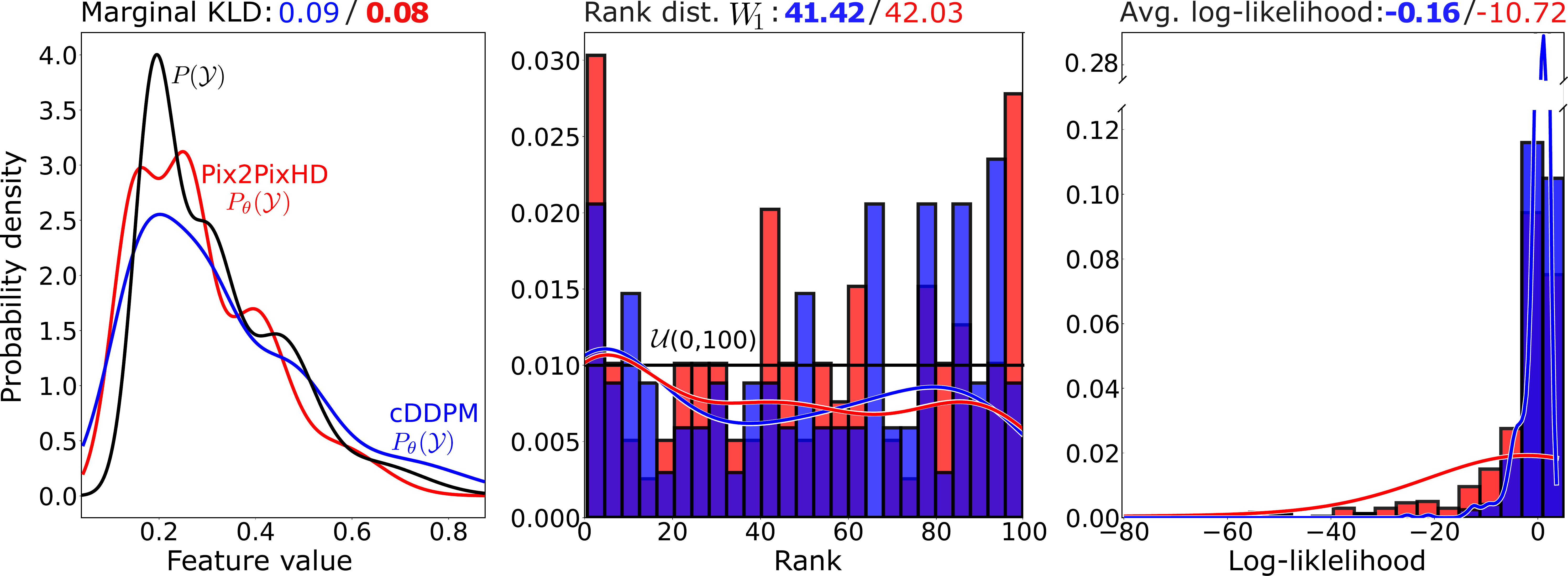}
    \caption{
    \textbf{Evaluation metrics for \emph{upper quartile intensity} (F7) feature.}
      Marginal KLD and rank distributions are inconclusive and mask differences in model performance. Log-likelihood distribution reveals superior performance of cDDPM.}
   
       
    \label{fig:contrastingresults}
        
    \end{figure}    
}

\newcommand{\figscatterpopvsmetrics}{%
\begin{figure}[t]
  \centering
  \includegraphics[width=\linewidth]{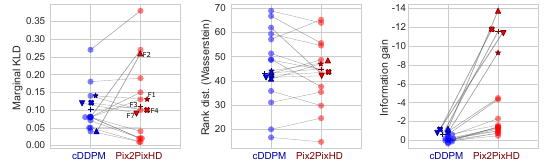}

  \vspace{4mm}

  {\fontsize{8}{9.5}\selectfont
  \setlength{\tabcolsep}{1pt}
  \renewcommand{\arraystretch}{1.1}

\begin{tabular}{llcccccccccc}
\hline
\textbf{Metric}                                                   & \textbf{Model} & \multicolumn{1}{c}{F1$\bigstar$} & \multicolumn{1}{c}{F2$\blacktriangle$} & \multicolumn{1}{c}{F3$\pmb{+}$} & \multicolumn{1}{c}{F4$\pmb{\times}$} & \multicolumn{1}{c}{F5} & \multicolumn{1}{c}{F6} & \multicolumn{1}{c}{F7 $\blacktriangledown$} & \multicolumn{1}{c}{F8} & \multicolumn{1}{c}{F9} & \multicolumn{1}{c}{F10-18} \\ \hline
\multirow{2}{*}{Marg. KLD ($\downarrow$)~}                   & cDDPM ~         & 0.14                   & \textit{0.04}          & \textit{0.10}          & 0.14                   & \textit{0.18}          & 0.12                   & 0.09                   & 0.08                   & 0.05                   & \textit{0.09}              \\
                                                                  & P2PHD          & \textit{0.13}          & 0.26                   & 0.11                   & \textit{0.10}          & 0.19                   & \textit{0.10}          & \textit{0.08}          & \textit{0.02}          & \textit{0.01}          & 0.12                       \\ \hline
\multirow{2}{*}{Rank $W_1$($\downarrow$)}                         & cDDPM          & \textit{44.12}         & \textit{43.02}         & \textit{43.05}         & \textit{42.20}         & 16.55                  & 35.77                  & \textit{41.42}         & 68.99                  & 66.80                  & 46.45                      \\
                                                                  & P2PHD          & 47.32                  & 48.66                  & 44.87                  & 43.58                  & \textit{14.67}         & \textit{29.68}         & 42.03                  & \textit{63.98}         & \textit{55.61}         & \textit{44.18}             \\ \hline
\multirow{2}{*}{Info gain ($\uparrow$)} & cDDPM          & \textbf{-0.82}         & \textbf{-1.22}         & \textbf{-0.67}         & \textbf{-1.17}         & \textbf{-0.94}         & \textbf{-0.64}         & \textbf{-0.80}         & \textbf{-0.01}         & \textbf{-0.01}         & \textbf{-0.1}              \\
                                                                  & P2PHD          & -9.29                  & -13.73                 & -11.52                 & -11.76                 & -4.48                  & -4.34                  & -11.36                 & -1.21                  & -0.75                  & -1.27                      \\ \hline
\end{tabular}
  }
  \caption{
  \textbf{Comparing evaluation metrics across features} (F1-F18). Metrics averaged for F10-F18 due to space limitations. Best model for each feature and metric is highlighted as bold/italic. 
  The proposed information gain shows a clear advantage for cDDPM across features, while the other metrics mask the performance difference.}
  \label{fig:swarm}
\end{figure}%
}
\title{A Proper Scoring Rule for Virtual Staining}
%
\author{
Samuel Tonks\inst{1}\orcidlink{https://orcid.org/0009-0008-1408-1493} \and
Steve Hood\inst{2}\orcidlink{https://orcid.org/0000-0002-7708-7699} \and
Ryan Musso\inst{3} \and
Ceridwen Hopely\inst{3} \and
Steve Titus\inst{3} \and
Minh Doan\inst{3}\orcidlink{https://orcid.org/0000-0002-3235-0457} \and
Iain Styles\inst{4}\orcidlink{https://orcid.org/0000-0002-6755-0299} \and
Alexander Krull\inst{1}\orcidlink{https://orcid.org/0000-0002-7778-7169}
}

\authorrunning{S.~Tonks et al.}

\institute{
School of Computer Science, University of Birmingham, Birmingham, UK\\
\email{smtonks2712@gmail.com, a.f.f.krull@bham.ac.uk}
\and
GSK Drug Metabolism \& Pharmacokinetics, 
GSK Medicines Research Centre, Stevenage, UK\\
\email{steve.r.hood@gsk.com}
\and
GSK Genome Biology, Collegeville, PA, USA\\
\email{ryan.x.musso@gsk.com, ceridwen.hopely@gsk.com, steve.titus@thermofisher.com, minh.x.doan@gsk.com}
\and
School of Electronics, Electrical Engineering and Computer Science, 
Queen's University Belfast, UK\\
\email{i.styles@qub.ac.uk}
}

  
\maketitle              
\begin{abstract}
Generative virtual staining (VS) models for high-throughput screening (HTS) can provide an estimated posterior distribution $P_{\theta}(\mathcal{Y}| \textbf{x})$ of possible biological feature values $\mathcal{Y}$ for each input and cell.
However, when evaluating a VS model, the true posterior $P(\mathcal{Y}| \textbf{x})$ is unavailable. 
Existing evaluation protocols only check the accuracy of the marginal distribution $P(\mathcal{Y})$ over the dataset rather than the predicted posteriors. 
We introduce \emph{information gain} (IG) as a cell-wise evaluation framework that enables direct assessment of predicted posteriors. 
IG is a \emph{strictly proper scoring rule} and comes with a sound theoretical motivation allowing for interpretability, and for comparing results across models and features. 
We evaluate diffusion- and GAN-based models on an extensive HTS dataset using IG and other metrics and show that IG can reveal substantial performance differences other metrics cannot.


\end{abstract}
\section{Introduction}
\figexplainerillposed
High-throughput screening (HTS) plays a crucial role in drug discovery by simultaneously assessing the effects of a large number of candidate compounds on cell cultures~\cite{szymanski2011adaptation}. 
A widely used technique in HTS is fluorescence microscopy~\cite{bradbury2020contrast}. 
Recorded fluorescence microscopy images are processed by cell detection~\cite{stringer2021cellpose} and feature extraction algorithms~\cite{carpenter2006cellprofiler}.
Numerical features $\mathcal{Y}_{i,j}$ for each cell $j$ in each image $i$, 
are computed that quantify, for example, the size of the nucleus, or the average intensity of the fluorescence in a particular region of the cell (Figure~\ref{fig:figexplainerillposed}, top). 
HTS uses statistics on these features to quantify the drug effect of potentially useful compounds. Because it can quickly screen a large number of drugs for possible effects, HTS is widely used in the drug discovery process.

HTS staining and imaging are time-consuming, costly, and require specialised equipment~\cite{ounkomol2018label}. 
Once fluorescent staining is applied, longitudinal observation is no longer possible, necessitating separate plates for each incubation time point, which introduces plate-specific distribution shifts known as plate effects~\cite{ding2017analysis}. 
To address these challenges, virtual staining (VS)~\cite{tonks2023evaluation,tonks2024,christiansen2018silico,ounkomol2018label,Tonks2025} has been successfully adopted as a cost-effective and scalable alternative.
Once trained on a dataset of paired brightfield images $\mathbf{x}_i$ and fluorescence images $\mathbf{y}_i$, VS models can translate new brightfield images into virtual fluorescence images, which can then be processed using a conventional HTS pipeline, see Figure~\ref{fig:figexplainerillposed} (bottom).

Early works~\cite{ounkomol2018label,christiansen2018silico} approached VS as a regression task, mapping brightfield images $\mathbf{x}$ to fluorescence predictions $\hat{\mathbf{y}} \approx \mathbf{y}$ via machine learning.
Such methods ignore the inherent uncertainty of the task: for an input brightfield image $\mathbf{x_i}$, it may not be possible to predict a single true fluorescent image $\mathbf{y}_i$.
From a Bayesian perspective, we describe this as a posterior distribution $P(\mathbf{y}_i | \mathbf{x}_i)$ of possible fluorescence images $\mathbf{y}_i$ and, in turn, of possible cell feature values $P(\mathcal{Y}_{i,j} | \mathbf{x}_{i,j})$.

More recently, conditional generative models have been used for VS~\cite{tonks2024,ho2020denoising,wang2018high}. 
They learn an approximate posterior distribution $P_\theta(\mathbf{y} | \mathbf{x})$ which can produce highly realistic and diverse samples.
A sampled virtual fluorescence image $\hat{\mathbf{y}}_i^k$ can be passed into cell detection and feature extraction algorithms, yielding a virtual feature value $\hat{\mathcal{Y}}_{i,j}^k$.
Repeated sampling and feature extraction enable us to derive an approximate posterior distribution of possible feature values $P_\theta(\mathcal{Y}_{i,j}|\textbf{x}_{i,j})$ for individual cells $j$ within an input image $\textbf{x}_i$.
The process is illustrated in Figure~\ref{fig:figexplainerillposed} (bottom).
In Figure~\ref{fig:metrics_deepdive}, we show sampled virtual fluorescence images and feature distributions from a single input image.

Recent years have seen a plethora of generative image-to-image approaches that could potentially be useful in VS.
Two popular choices are generative adversarial networks (GANs)~\cite{GenerativeAdversarialNets,wang2018high,brock2018large,upadhyay2021uncertainty} and diffusion models  (DMs)~\cite{dhariwal2021diffusion,ho2020denoising,batzolis2021conditional,saharia2022palette,peebles2023scalable}.
While GANs allow for fast sampling, they often suffer from mode collapse~\cite{GenerativeAdversarialNets} and inferior sample quality~\cite{dhariwal2021diffusion}.
DMs are known to produce more realistic samples and better distribution coverage, but require more time and computational cost for sampling~\cite{ho2020denoising,dhariwal2021diffusion}.
Assessing and comparing the quality of such methods in the VS context is essential for their evidence-based application, but it is far from obvious what the correct evaluation metric should be.

Sample quality and coverage in generative image models are often evaluated using image-based metrics such as the Inception Score (IS)~\cite{barratt2018note}, Fréchet Inception Distance (FID)~\cite{preuer2018frechet,soloveitchik2021conditional}.
Such methods use a pre-trained neural network to map images to a latent space and then compare the distributions of real and sampled data.
However, the learned latent space is usually trained from natural images, and is not necessarily meaningful for an application such as HTS.
An arguably more meaningful approach is evaluation at the feature level and many works~\cite{tonks2023evaluation,christiansen2018silico,ounkomol2018label,saharia2022image,razghandi2022variational} evaluate model performance by aggregating feature samples $\hat{\mathcal{Y}}_{i,j}$ across cells, comparing the result to the marginal feature distribution $P(\mathcal{Y})$ across the dataset. 
We show such a comparison in Figure~\ref{fig:contrastingresults} (left).
However, while a perfect model must lead to a perfect approximation of the true feature marginal, this approach does not assess the accuracy of the individual cell posteriors.

\indent We propose that a more natural and meaningful approach would instead attempt to directly evaluate the predicted posterior distributions of image features $P_\theta(\mathcal{Y}_{i,j}|\textbf{x}_{i,j})$ for each cell by comparing it to the true posterior $P(\mathcal{Y}_{i,j}|\textbf{x}_{i,j})$ of possible feature values for each given input $\textbf{x}_{i,j}$.
Such an approach would allow us to assess how well a model can extract information about the feature from the corresponding brightfield image.
However, while the probability density function (PDF) $P_\theta(\mathcal{Y}_{i,j}|\textbf{x}_{i,j})$ can be approximated through sampling and a kernel density estimator (KDE) or by fitting a Gaussian mixture model (GMM), the true posterior PDF $P(\mathcal{Y}_{i,j}|\textbf{x}_{i,j})$ for each brightfield image $\textbf{x}_i$ is unavailable.
For each brightfield input $\mathbf{x}_i$, only a single fluorescence image $\textbf{y}_i$ and thus only a single feature value $\mathcal{Y}_{i,j}$, i.e. a single sample from $P(\mathcal{Y}_{i,j}|\mathbf{x}_{i,j})$ is available, see Figure~\ref{fig:figexplainerillposed} (right).
This raises a key question: \textit{how can we assess the quality of the predicted posterior $P_\theta(\mathcal{Y}_{i,j}|\mathbf{x}_{i,j})$ when only a single sample from the true posterior is available?}

Similar problems in statistical forecasting for metrology and other fields are regularly addressed via \emph{scoring rules}~\cite{gneiting2007strictly}.
However, in generative image modelling and in particular VS, these approaches are largely unknown and feature-level evaluation metrics have not been systematically explored.
In this work, we address this gap and consider three evaluation metrics designed to assess posterior quality at the cellular level, even when only a single true sample is available.
These are based on:
$(i)$ Comparing the predicted and real marginal feature distribution,
$(ii)$ the ranks of the true feature values in the context of the predicted distribution, and $(iii)$ the gain of the average log-probability of the true value under the predicted posterior compared to the marginal (also known as information gain (IG)). 

We use a 30k image VS dataset to systematically compare marginal-level metric $(i)$ and posterior-level metrics $(ii, iii)$ against each other using DM-and GAN-based VS models. 
Our results demonstrate that evaluation metrics $(i)$ and $(ii)$ mask substantial differences in posterior quality. 
In contrast, $(iii)$, which is a \emph{strictly proper scoring rule}~\cite{gneiting2007strictly}, can reveal substantial performance differences between models and among different features.

\section{Methods} \label{methods}
Consider a generative VS model with learnt parameters $\theta$.
When presented with a brightfield image $\textbf{x}_i$, the model can produce samples $\hat{\textbf{y}}_i^k$ from an approximate posterior $P_\theta(\textbf{y}_i|\textbf{x}_i)$ and by applying the feature extraction pipeline (Figure~\ref{fig:figexplainerillposed} bottom) we can derive sampled individual cell feature values $\hat{\mathcal{Y}}^k_{i,j}$ from an implicit posterior $P_\theta(\mathcal{Y}_{i,j}|\textbf{x}_{i,j})$.
Our goal is now to use an evaluation dataset of pairs of $(\textbf{x}_{i,j}, \mathcal{Y}_{i,j})$ to quantify how well the predicted posterior $P_\theta(\mathcal{Y}_{i,j}|\textbf{x}_{i,j})$ fits the unknown true distributions $P(\mathcal{Y}_{i,j}|\textbf{x}_{i,j})$ and to aggregate this measure over our dataset in order to allow for an assessment of the overall model performance.

\noindent \textbf{The marginal KLD metric} circumvents the problem that the true posterior is unknown and instead compares the feature marginal $P(\mathcal{Y})$ with a marginal $P_\theta(\mathcal{Y})$ derived from the model.
The probability density $P(\mathcal{Y})$ can be approximated by fitting a GMM or KDE to the combined true feature values $\mathcal{Y}_{i,j}$. 
The model marginal $P_\theta(\mathcal{Y})$ can be approximated in the same way by using sampled feature values  $\mathcal{Y}^k_{i,j}$ from all cells across the dataset.
The resulting metric is defined as 
\begin{equation}
\text{Marginal KLD} = D_{\mathrm{KL}}\left(P(\mathcal{Y})\| P_\theta(\mathcal{Y})\right),
\end{equation}
where $D_{\mathrm{KL}}(\cdot\|\cdot)$ is the Kullback-Leibler divergence (KLD).

Rather than avoiding the problem by focusing on the marginal, the remaining two evaluation metrics directly evaluate the quality of the predicted posteriors.

\noindent \textbf{The rank metric}, or probability integral transform~\cite{gneiting2007probabilistic,diebold1997evaluating}, 
is based on the rank order $r_{i,j}$ of the true feature value $\mathcal{Y}_{i,j}$ among $K$ samples $\hat{\mathcal{Y}}_{i,j}^k \sim P_\theta(\mathcal{Y}_{i,j}| \mathbf{x}_{i,j})$, that is, the number of samples smaller than $\mathcal{Y}_{i,j}$.
If $P_\theta(\mathcal{Y}_{i,j}|\mathbf{x}_{i,j})$ is a perfect model of the true posterior $P(\mathcal{Y}_{i,j}| \mathbf{x}_{i,j})$ for all $\mathbf{x}_{i,j}$
then the collection of $r_{i,j}$ combined over all cells should be uniformly distributed over $\{0,\dots,K\}$.
We define the metric as the deviation from uniformity using 
\begin{equation}
\text{Rank Distance} =W_1\!\left(P_\theta(r),\,\mathcal{U}(0,100)\right),  
\end{equation}
where $W_1(\cdot, \cdot)$ is the 1-Wasserstein distance~\cite{ruschendorf1985wasserstein} and $P_\theta(r)$ denotes the empirical distribution of ranks $\{r_{i,j}\}$ normalized between 0 and 100 over all cells $(i,j)$. The rank distributions for a single feature are shown in Figure~\ref{fig:contrastingresults}.

\noindent \textbf{The information gain metric} is based on the average log-likelihood
\begin{equation}
  \bar{\ell} = 
 \frac{1}{N}
 \sum_{i=1}^{M}
 \sum\limits_{j=1}^{N_i}
 \log P_\theta(\mathcal{Y}_{i,j}|\textbf{x}_{i,j}),
  \label{eq:log_like}
\end{equation}
where \(P_\theta(\mathcal{Y}_{i,j}|\textbf{x}_{i,j})\) is the density estimation of the predicted biological feature posterior distribution $P(\mathcal{Y}_{i,j}|\textbf{x}_{i,j})$. 
We denote the number images as $M$, the number of cells in an image $i$ as $N_i$ and the total number of cells as $N$.

For large $N$, $\bar{\ell}_\theta$ approximates the average negative Kullback-Leibler divergence  (KLD) over the dataset up to a constant shift~\cite{gneiting2007strictly,Tonks2025}:
\begin{equation}
\bar{\ell}_\theta\approx
  -\frac{1}{N}
 \sum_{i=1}^{M}
 \sum\limits_{j=1}^{N_i}
  D_{\mathrm{KL}}\left(P(\mathcal{Y}_{i,j}|\textbf{x}_{i,j})\| P_\theta(\mathcal{Y}_{i,j}|\textbf{x}_{i,j})\right) + \text{constant}.
  \label{eq:all_kl}
\end{equation}
In practice, \(P(\mathcal{Y}_{i,j}|\textbf{x}_{i,j})\) and \(P_\theta(\mathcal{Y}_{i,j}|\textbf{x}_{i,j})\) could again be any KDE or GMM fitted to sampled feature values $\hat{\mathcal{Y}}_{i,j}^k$.
Eq.~\ref{eq:log_like} corresponds to the logarithmic score~\cite{gneiting2007strictly}, which is a strictly proper scoring rule, meaning the value is maximised iff predicted distributions $P_\theta(\mathcal{Y}_{i,j}|\textbf{x}_{i,j})$ are equal to the true posteriors $P(\mathcal{Y}_{i,j}|\textbf{x}_{i,j})$.

While we can use Eq.~\ref{eq:log_like} to compare two models in terms of how well they approximate the true posteriors, the unknown $constant$ depends on the data/feature and prevents $\bar{\ell}$ from directly reflecting the quality of a model.  
Thus, 
we cannot use $\bar{\ell}_\theta$ to compare the quality of predictions between different features or datasets.
To overcome this, we compute the logarithmic score of our model relative to a simple reference model.
This is known as \emph{information gain score}, which is still a strictly proper scoring rule~\cite{peirolo2011information,takaya2023information}.
We use the feature marginal $P(\mathcal{Y})$ as the reference model and compute $\bar\ell_{\text{ref}}$ analogous to Eq.~\ref{eq:log_like} as the average log-likelihood according to the marginal. 
The resulting score is
\begin{equation}
 \text{Info gain} =  \bar{\ell}_\theta - \bar\ell_\text{ref}
.
\label{eq:ig}
\end{equation}
Eq.~\ref{eq:ig} leads to the cancellation of the constant from Eq.~\ref{eq:all_kl}, yielding a metric that now allows not only for the direct comparison between models but also across features and datasets.
This can be viewed as the reduction in the average KLD (see Eq.~\ref{eq:all_kl}) achieved by the model over the marginal or as the amount of information the model is able to extract from the brightfield image~\cite{peirolo2011information}. 
\metricsdd
\section{Experiments \& Results}  \label{experiments}
In this section, we evaluate two generative VS models to assess the insights provided by our proposed metrics.

\noindent \textbf{Dataset:} We used the dataset made available to us by the authors of~\cite{tonks2024}.
The dataset contains 30,000 images generated for an HTS assay. 
Brightfield and DAPI image pairs were randomly selected from 49 16x24 well plates, each containing negative and positive controls, and 10 compounds screened for toxicity for ovarian cell types from six different cell lines. 
Brightfield and fluorescence stain image pairs were randomly split (70/20/10).
Corresponding nuclei masks were obtained using a pre-trained nuclei segmentation model from Cellpose~\cite{stringer2021cellpose}, a gamma correction of 0.3 was applied. 
The feature extraction pipeline shown in Figure~\ref{fig:figexplainerillposed} follows the same protocol as in~\cite{tonks2024} to obtain cell-wise biological feature values using the DAPI image and Cellpose mask pairs. We use the nuclei intensity features 'LowerQuartile', 'MeanEdge','Mean', 'Median', 'StdEdge', 'Std' and 'UpperQuartile' (F1-F7) and all radial distribution features from categories 'FracAtD', 'MeanFrac' and 'RadialCV' (F8-F18)~\cite{cimini2023cellpainting}.

\noindent \textbf{Training \& Inference:}
This work evaluated our proposed metrics using Pix2pix-HD GAN~\cite{wang2018high} and Conditional Denoising Diffusion Probabilistic Model (cDDPM)~\cite{ho2020denoising}.
Pix2pixHD used hyper-parameters from ~\cite{wang2018high} but with three discriminators operating at 256, 512 and 1024 pixel resolutions.
cDDPM, an adaptation of \cite{ho2020denoising} in which the generator takes the brightfield as an additional input channel, was trained on random 256x256 image crops using Adam~\cite{kingma2014adam}, learning rate $\mathrm{1e{-}5}$ and an exponential moving average decay rate.
Both models were trained for a maximum of 200 epoch with early stopping. 
During inference, 1,000 virtual nuclei samples were obtained for each brightfield input from each model.
MC-Dropout~\cite{gal2016dropout} was used in Pix2pixHD GAN to enable sampling.

\noindent \textbf{Per-cell image and feature distributions:}
To illustrate the visual diversity of outputs of cDDPM and Pix2pixHD, we show multiple sampled virtual images from both models for a single input in Figure~\ref{fig:metrics_deepdive} (left).
Qualitatively, we observe that cDDPM is better able to reproduce the true shape, scale and intensity profile of the cell compared to Pix2pixHD. 
Additionally, cDDPM produces much more variability in samples compared to Pix2pixHD. 
We generated 1,000 samples for each model and computed the corresponding values of the 18 cell-profiler features.
The resulting feature distribution for the \emph{upper quartile intensity} (F7) feature is shown in Figure~\ref{fig:metrics_deepdive} (right).
We find that the corresponding cell-wise posterior distribution for cDDPM is substantially wider and more centred on the target value compared to Pix2pixHD.

\figcmodels

\noindent \textbf{Aggregated quality measures:}
In Figure~\ref{fig:contrastingresults} (left), we show GMM-based density estimates across all cells for both models compared to the target for the F7 feature.
Using our marginal KLD approach we compare the true and predicted marginals.
Visually the results are inconclusive with both models fitting about equally well and achieving similar KLDs.
Examining the centre plot containing the rank distributions, for Pix2pixHD, we observe 
much like for the marginal KLD, the rank distributions do not give a clear picture and suggest similar performance of both models.
\figscatterpopvsmetrics
The right plot of Figure~\ref{fig:contrastingresults} shows the distribution of log-likelihoods over the cells.
Pix2pixHD has a considerably longer left tail compared to cDDPM, suggesting that for many cells, the probability of the true sample value occurring within the predicted posterior is very low.
The highlighted average log-likelihood (Eq.~\ref{eq:log_like}) of cDDPM is $10.54$ higher than Pix2pixHD, despite Pix2pixHD achieving a lower marginal KLD.

 

We see a clear difference in the marginal KLD compared to the log-likelihood distributions suggesting that Pix2pixHD is predicting realistic feature values but not for the right cells and is thus not correctly utilising the input signal.

When comparing the rank distance and log-likelihood there is an apparent contradiction;
where the rank distribution indicates a similar posterior quality, the log-likelihood shows a clear difference between the models.
Since the average log-likelihood is a strictly proper scoring rule and the rank distance is not, we believe this is a real difference, which is not picked up by he rank distance.
This conclusion is corroborated by the qualitative results from Figure~\ref{fig:metrics_deepdive}.

\noindent \textbf{Comparing overall performance across models and features:}
To assess the performance of the two trained models, we calculate the marginal KLD as well as our proposed two metrics for \emph{all features}.
Figure~\ref{fig:swarm} shows the results obtained for cDDPM and Pix2pixHD, both within and across metrics.

Comparing the marginal KLD with the rank distance, there is no clear trend of features performing differently or one method consistently outperforming the other.
However, complementary to Figures~\ref{fig:metrics_deepdive} and \ref{fig:contrastingresults} when using IG, cDDPM outperforms Pix2pixHD across all features.
Additionally, we can see a cluster of highlighted features (F1,F3,F4,F7) for which Pix2pixHD dramatically underperforms even compared to other features.
In contrast, cDDPM's results are highly concentrated and show no outlier features.
The previously explored F7 feature in Figure~\ref{fig:contrastingresults} is part of this cluster.
Notably, all of these underperforming features are intensity-based (F1-F9), revealing an apparent insufficiency in the method to correctly extract intensity information from brightfield data.
This cluster remains completely hidden in the other two metrics, where the relevant features are positioned in amongst the other features for Pix2pixHD.

In the IG metric 
results above 0 indicate the model is performing better than the marginal feature distribution in describing the posterior.
None of the Pix2pixHD results and only 5 of the cDDPM results are above zero, indicating that even though cDDPM manages to produce visually realistic results, there is substantial room for improvement.

\section{Conclusion}
We propose information gain (IG), a strictly proper scoring rule, to directly evaluate the predicted posterior distributions of biological features in virtual staining for HTS.
Empirical evaluation across two VS approaches (Pix2pixHD and cDDPM) demonstrates substantial differences in posterior accuracy and sample variability that are captured by IG but not by marginal KLD or rank distance.
While both metrics suggest VS models have similar performance, our scoring rule finds dramatic differences in performance for many features. 
This indicates that models such as Pix2pixHD may recover approximately correct marginal distributions but fail to produce input-consistent conditional predictions for individual samples, leading to overestimated performance when evaluated using marginal metrics alone.
It is clear that the marginal KLD for VS is not aligned with the conditional modelling objective, as it assesses performance with respect to the marginal distribution $P(\mathcal{Y})$ rather than the cell posterior $P(\mathcal{Y}|\textbf{x})$.
However, also the rank distance did not conclusively identify incorrectly predicted posterior distributions.
Using these metrics may lead to systematic overestimation of model performance in conditional generative models tasks.
By explicitly quantifying the IG from the input signal, we provide a principled means of assessing posterior adherence in virtual staining.
We believe that the same procedure may be applicable for other image-to-image translation models in downstream predictive tasks beyond virtual staining or microscopy.

\section{Acknowledgments}
This work was funded by GSK and a studentship to ST from the Engineering and Physical Sciences Research Council (reference 2457518). For the purpose of Open Access, the author has applied a CC BY public copyright license to any Author Accepted Manuscript version arising from this submission. 

\bibliographystyle{splncs04}
\bibliography{main}
\end{document}